\documentclass[10pt,twocolumn,letterpaper]{article}

\usepackage[pagenumbers]{cvpr} % To force page numbers, e.g. for an arXiv version

\usepackage{multirow}
\usepackage{color}
\usepackage{colortbl}
\usepackage{amsmath}
\usepackage{tabularx}
\usepackage{adjustbox}
\usepackage{array}
\usepackage{booktabs}

\definecolor{cvprblue}{rgb}{0.21,0.49,0.74}
\usepackage[pagebackref,breaklinks,colorlinks,allcolors=cvprblue]{hyperref}

\def\paperID{22227} % *** Enter the Paper ID here
\def\confName{CVPR}
\def\confYear{2026}

\title{TransPrune: \\Token Transition Pruning for Efficient Large Vision-Language Model}

\author{
Ao Li\thanks{Equal contribution to this work.}\textsuperscript{\rm ~~,1} \quad 
Yuxiang Duan\footnotemark[1]\textsuperscript{\rm ~~,1} \quad 
Jinghui Zhang\textsuperscript{2} \quad
Congbo Ma\textsuperscript{3} \\
Yutong Xie\textsuperscript{2} \quad
Gustavo Carneiro\textsuperscript{4} \quad
Mohammad Yaqub\textsuperscript{2} \quad
Hu Wang\thanks{Corresponding author.}\textsuperscript{\rm ~~,2}
\\
\textsuperscript{1}Shandong University~% 
\textsuperscript{2}MBZUAI~%
\textsuperscript{3}New York University Abu Dhabi
\textsuperscript{4}University of Surrey~% 
}

\begin{document}
\maketitle

\begin{abstract}
Large Vision-Language Models (LVLMs) have advanced multimodal learning but face high computational cost issues due to the input of large number of visual tokens, motivating token pruning to improve inference efficiency.
The key challenge lies in identifying which tokens are truly important.
Most existing approaches rely on attention- or similarity-based criteria to estimate token importance.
However, they inherently suffer from certain limitations, such as being task-agnostic and exhibiting positional bias.
In this work, we explore a new perspective on token importance assignment based on token transitions in LVLMs, where token transitions are defined as the changes in token representations occurring as they propagate through the model’s modules.
We observe that \textbf{the transition of token representations provides a meaningful signal of semantic information.} 
Based on this insight, we propose \textbf{TransPrune}, a training-free and efficient token pruning method.
Specifically, TransPrune progressively prunes tokens by assessing their importance through a combination of Token Transition Variation (TTV), which measures changes in both the magnitude and direction of token representations; as well as Instruction-Guided Attention (IGA), which measures how strongly the instruction attends to visual tokens via attention.
Extensive experiments on various LVLM architectures, such as LLaVA-v1.5, LLaVA-Next and Qwen2.5-VL, demonstrate that TransPrune maintains comparable multimodal performance while reducing inference TFLOPs by more than half.
\end{abstract}

\section{Introduction}

\begin{figure}[!ht]
    \centering
    \includegraphics[width=\linewidth]{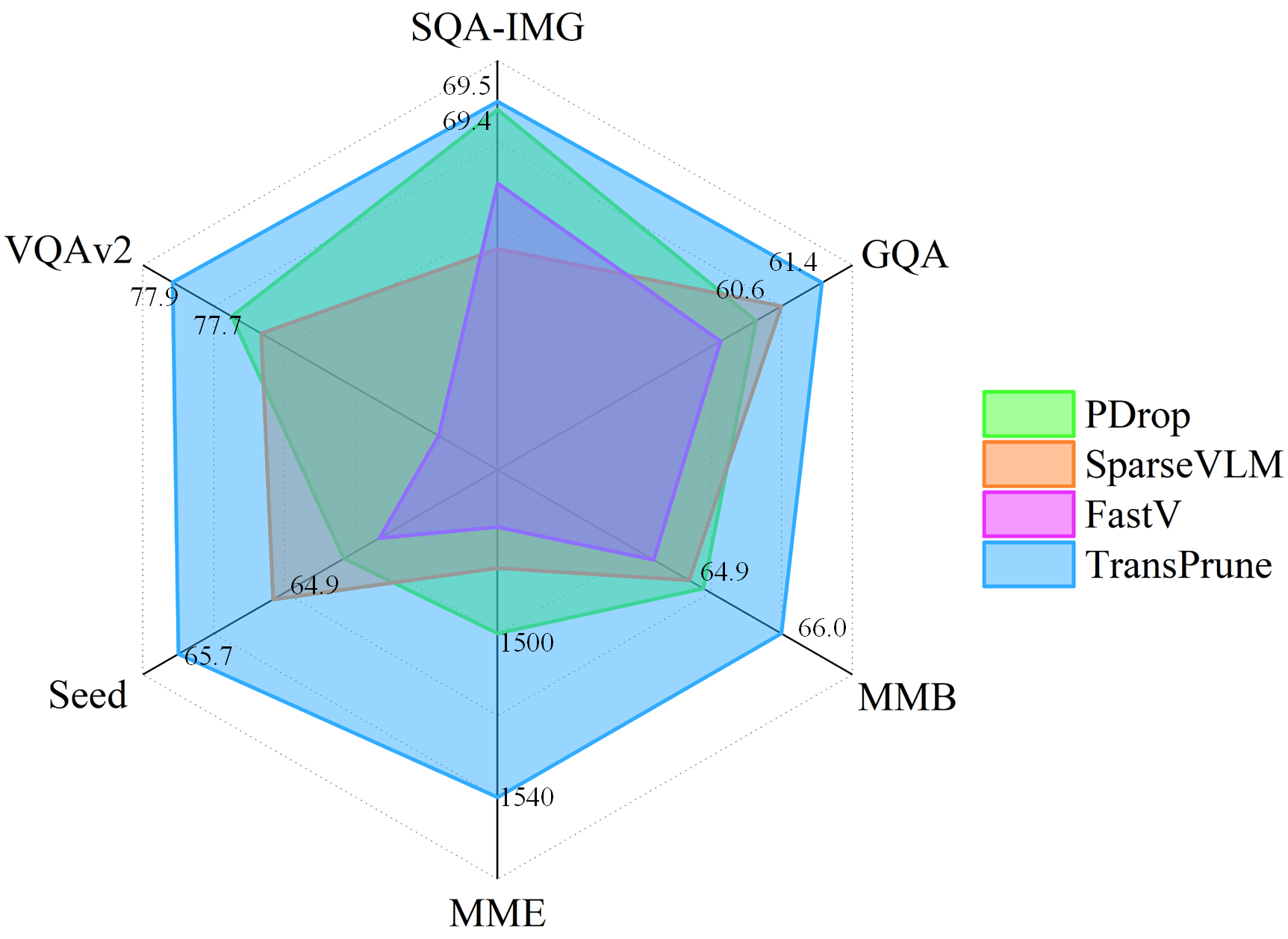}
    \caption{Comparison with existing pruning methods on LLaVA-v1.5-7B. Among within-LLM pruning approaches, TransPrune achieves the best performance across six benchmarks under the lowest TFLOPs budget.}
    \label{fig:performance}
\end{figure}

\begin{figure*}[!ht]
    \centering
    \includegraphics[width=\linewidth]{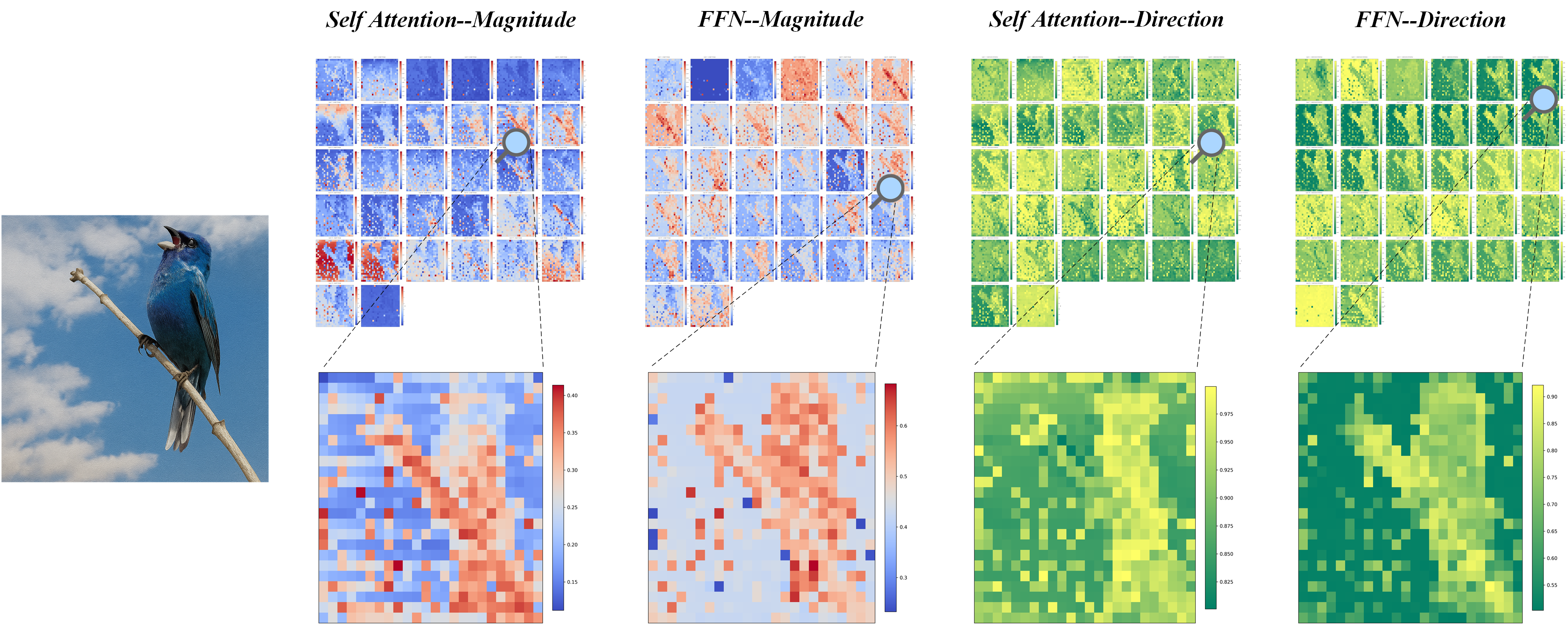}
    \caption{Token Transition Visualization in LLaVA-v1.5-7B.
    We visualize the magnitude and direction changes of token representations within both the self-attention and FFN modules for each layer (excluding residual connections).  
    To measure the magnitude change, we use the ratio of output to input L2 norm; to measure the directional change, we use cosine similarity.
    Token transitions that reflect semantic importance can be observed across shallow, middle, and deep layers, and they are most concentrated and pronounced in the middle layers (around layers 6–14), where tokens with larger ratios and smaller absolute cosine similarities tend to be more semantically important.
    We provide more visualization examples in supplementary material. 
    } 
    % (i.e., output tending more orthogonal to input) 
    \label{fig:vis}
\end{figure*}

Recently, Large Vision-Language Models (LVLMs) have achieved remarkable progress, demonstrating impressive performance on a wide range of tasks~\cite{MiniGPT4,BLIP2}.
However, LVLMs typically incur substantial computational overhead.
A primary contributor to this computational burden is the large number of visual tokens processed during inference.
Consequently, an effective way to improve the efficiency of LVLMs is token pruning, which identifies and retains the most important tokens that typically carry richer semantic information and are more closely related to the user’s instruction.
However, reliably estimating the importance of each visual token remains a challenging problem.

Token pruning methods are generally categorized into \textbf{\textit{within-LLM}}~\cite{fastv,pdrop,sparsevlm} and \textbf{\textit{projector-based}} approaches~\cite{visionzip,cdpruner,vispruner,DivPrune}, all of which fundamentally rely on attention or similarity-based criteria to identify the most informative tokens.
While attention-based methods are widely used and often effective, they exhibit inherent limitations.
In particular, attention exhibits positional bias~\cite{pact,problem}, where tokens at the beginning or end of a sequence are often assigned higher attention scores compared to those in other positions.
Additionally, attention may overemphasize visually salient but semantically irrelevant regions~\cite{attention}.
% Consequently, relying solely on attention to assess token importance may result in a suboptimal evaluation.
In contrast, similarity-based methods typically merge tokens with high representational similarity, making them task-agnostic and less effective at identifying tokens that are truly relevant to specific downstream tasks.
These limitations motivate the exploration of alternative or complementary criteria for token importance estimation.

In numerous real-world phenomena, the dynamic evolution and transformation of an entity often yield deeper and more nuanced insights than a mere examination of its instantaneous or static state~\cite{dynamicvit}. 
Inspired by this broader perspective, we delve into a novel viewpoint: \textit{\textbf{can the dynamic transition of a token representation serve as an indicator of its importance?}}

To answer this question, we evaluate token transitions from two complementary perspectives: first, the magnitude change, quantified by the L2 norm between a token's input and output representations within a module; and second, the direction change, captured by the cosine similarity between these two representations~\cite{transf}.
We visualize token transitions across different layers, focusing on both the self-attention and feed-forward network (FFN) modules, as shown in Figure~\ref{fig:vis}.
Interestingly, transitions in the middle layers indeed reflect the semantic information of tokens.

Based on our observation, we propose \textbf{TransPrune}, which primarily leverages two complementary criteria to estimate token importance: Token Transition Variation (TTV) and Instruction-Guided Attention (IGA).
TTV captures both the magnitude and direction changes of tokens representations by focusing solely on each token’s self-transition, without computing inter-token dependencies. 
This design avoids the positional bias that may arise from the triangular mask mechanism in attention~\cite{pact}.
Complementing TTV, IGA estimates token importance based on attention scores from instruction tokens, introducing task-guided semantic supervision. 
% Together, these two signals provide a robust foundation for pruning uninformative tokens.
However, while TTV reflects token importance, its patterns are not consistently stable across layers.
To mitigate this variability, we propose an accumulation mechanism that aggregates TTV values exclusively across middle layers exhibiting consistent characteristics, thereby yielding a more reliable importance metric.

Extensive experiments demonstrate that TransPrune, as an effective within-LLM pruning method, achieves comparable multimodal performance with over 50\% TFLOPs reduction.
Furthermore, combining TransPrune with projector-based methods, such as VisionZip~\cite{visionzip} and CDPruner~\cite{cdpruner}, can further improve token reduction efficiency while maintaining multimodal performance.

In summary, our main contributions are as follows:
\begin{itemize}
    \item We introduce a novel perspective beyond attention and similarity mechanisms by showing that token transitions yield meaningful signals of token importance in LVLMs.
    \item We propose TransPrune, which combines TTV and IGA. 
    TTV captures magnitude and direction changes of visual tokens, while IGA estimates image token importance based on attention from instruction tokens.
    \item Extensive experiments show that TransPrune maintains comparable multimodal performance while reducing inference TFLOPs by more than half on LLaVA-v1.5, LLaVA-Next and Qwen2.5-VL. 

\end{itemize}
\section{Related Work}

\subsection{Large Vision-Language Models} 
Large Vision-Language Models (LVLMs) have achieved remarkable progress in multimodal comprehension and generation~\cite{gpt4v,gemini,llava-1.5,MiniGPT4,BLIP2}.
Representative models such as LLaVA~\cite{llava-1.5}, BLIP2~\cite{BLIP2}, and MiniGPT-4~\cite{MiniGPT4} enable users to interact with the system through rich multimodal prompts that encompass both textual and visual inputs.
Beyond general-purpose tasks, recent advancements have further extended the capabilities of LVLMs to downstream applications such as affective computing~\cite{emoverse,zhang2025individuals} and medical image understanding~\cite{m3d}, demonstrating their potential in high-stakes, domain-specific scenarios.

Despite recent advances, the inherent complexity of LVLMs still demands substantial computation for both training and inference. 
This challenge intensifies with fine-grained understanding of high-resolution images and becomes even more severe for video understanding~\cite{video-chatgpt,video-llava,video-xl-pro}, where temporal redundancy and long sequences drastically increase token counts. 
These growing burdens highlight the urgent need for efficient token pruning to enable scalable, real-time LVLMs.

\subsection{Token Pruning}
The existing token pruning methods for LVLMs can be categorized into two types: \textbf{projection-based pruning~\cite{visionzip, cdpruner,vispruner,DivPrune,gridprune} and within-LLM pruning~\cite{fastv,pdrop,sparsevlm,topv,shortv}.}

Projector-based pruning methods are designed to select and prune visual tokens before passing them to the LLM.
VisionZip~\cite{visionzip} selects frequently attended visual tokens and merges similar ones based on their similarity to reduce redundancy.
DivPrune~\cite{DivPrune} formulates pruning as a Max-Min Diversity Problem to select the most diverse subset of tokens, thereby reducing redundancy while preserving performance.
CDPruner~\cite{cdpruner} leverages conditional diversity to select tokens based on user instructions.

Although projector-based methods are receiving increasing attention, exploring within-LLM pruning approaches remains essential, as they can leverage information unique to the LLM that is not captured by the visual encoder. 
Moreover, combining these two types of methods can further improve inference efficiency. 
Specifically, within-LLM pruning operates within the LLM itself. Such methods typically perform token pruning at different layers based on internal evaluation metrics of tokens within the LLM.
FastV~\cite{fastv} prunes visual tokens based on their attention from the last token. 
PDrop~\cite{pdrop} adopts a pyramid-style multi-stage pruning strategy to accelerate inference. 
SparseVLM~\cite{sparsevlm} leverages attention scores between important instruction tokens and visual tokens to guide pruning.
However, recent studies have highlighted that reliance on attention scores alone can introduce positional bias~\cite{pact,problem}.

Unlike most existing within-LLM approaches that rely solely on attention or similarity, TransPrune introduce a novel and efficient criterion grounded in token transition to evaluate token importance better.

\begin{figure*}[!ht]
    \centering
    \includegraphics[width=1.0\linewidth]{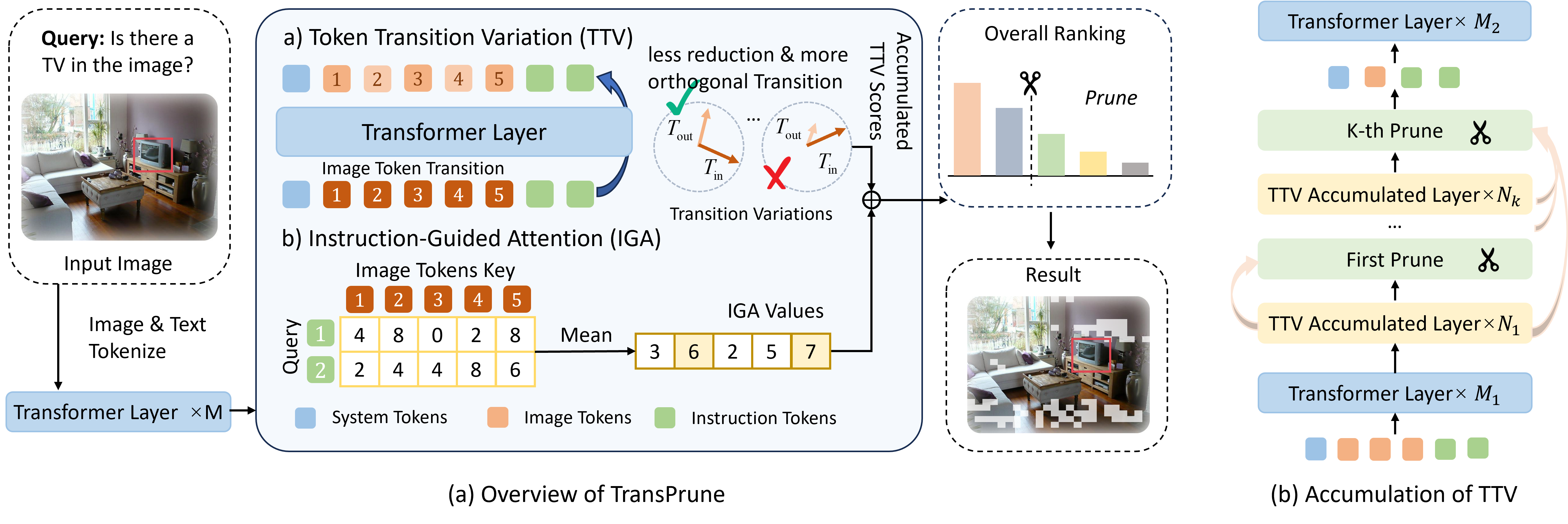}
    \caption{
    (a) Overview of TransPrune.  
    During pruning, TransPrune computes image token transitions.  
    Tokens whose transitions are closer in magnitude to those of the original tokens, and that exhibit more orthogonal directional changes, are assigned higher TTV scores.  
    In parallel, we compute IGA by averaging the attention from instruction tokens to image tokens.  
    The final score for each token is obtained by summing TTV and IGA, followed by sorting.
    (b) Accumulation of TTV.  
    To achieve a more precise TTV, we retain TTV scores from earlier layers.  
    For each pruning stage, we accumulate TTV scores from the first accumulated layer up to the current pruning layer.
    } 
    \label{fig:method}
\end{figure*}

\section{Method}
In this section, we introduce the TransPrune method, as shown in Figure~\ref{fig:method}.
\subsection{Token Transition Variation}

\noindent{\textbf{Token transition as a signal of importance.}}
Inspired by the insight that dynamic changes often better reflect the state of an entity than static values in the real world, we hypothesize that the transformations experienced by tokens within LLM layers may reveal their semantic importance. 

Each transformer layer consists of a self-attention module and a FFN, both of which modify token representations in meaningful ways.
To characterize these changes, we analyze token transitions along two dimensions: \emph{magnitude} and \emph{direction}. 
Formally, let $F$ denote a transformation module (e.g., self-attention or FFN), and let $T_{\text{in}}$ be the input token representation. 
The output token is then given by $T_{\text{out}}=F(T_{\text{in}})$. 
We define the magnitude transition $m(F, T_{\text{in}})$ and direction transition $d(F, T_{\text{in}})$ as:
\begin{equation}\small
    m(F, T_{\text{in}}) = \frac{\left\| T_{\text{out}} \right\|_2}{\left\| T_{\text{in}} \right\|_2}, 
    d(F, T_{\text{in}}) = \frac{T_{\text{out}} \cdot T_{\text{in}}}{\left\| T_{\text{out}} \right\|_2 \left\| T_{\text{in}} \right\|_2}.
\end{equation}
We empirically observe (see Figure~\ref{fig:vis}) that these two types of transition variations reflect a token's semantic information, which can be observed to varying degrees across shallow, middle, and deep layers, but is most pronounced and concentrated in the middle layers.
We argue that this phenomenon arises because the middle layers lie between the shallow global features and the deep local features~\cite{fastv}, enabling them to integrate both global context and local details. 
As a result, \textbf{these layer-wise transitions demonstrate how the LLM progressively shifts its attention from global representations to local information under the guidance of instructions.}
Consequently, tokens exhibiting larger transitions in the middle layers better reflect the model’s dynamic attention shifts and are therefore more semantically important.

Based on this insight, we propose a criterion called \textbf{Token Transition Variation (TTV)}. 
Specifically, we find that $1 - \left| d(F, T_{I}) \right|$ performs better than $d(F, T_{I})$ directly. 
Accordingly, we define the direction transition of all image tokens $T_{I}$ using $1 - \left| d(F, T_{I}) \right|$ (see the supplementary material).
We then apply a softmax operation across all tokens to normalize these direction transition values and multiply them by the corresponding magnitude transitions to compute the final TTV score:
\begin{equation}\small
\text{TTV}(F, T_{I}) = \text{Softmax}\left(1 - \left| d(F, T_{I}) \right| \right) \cdot m(F, T_{I}).
\end{equation}
For each layer $l$, we compute a token’s TTV score by aggregating the contributions from both the self-attention and FFN modules:
\begin{equation}\small
\text{TTV}_{l}(T_{I}) = \text{TTV}(Attention, T_{I}) + \text{TTV}(FFN, T_{I}).
\end{equation}
These scores are then used to guide token pruning decisions in each layer.

\noindent{\textbf{Token Transition Accumulation for Precise Pruning.}}
In token pruning, it is common to remove tokens at specific layers based on a predefined importance criterion.
However, as shown in Figure~\ref{fig:vis}, the TTV patterns vary across layers, making TTV scores from any single layer insufficiently precise for consistently identifying important tokens.
To improve the consistency of token pruning, we introduce an accumulation mechanism that aggregates token transitions across multiple layers before each pruning layer, as shown in Figure~\ref{fig:method} (b).

Formally, we define the accumulation layer set $\mathcal{A} = \{a_1, a_2, \dots, a_m\}$, where TTV scores are computed. 
Within this set $\mathcal{A}$, we select specific layers, forming the pruning layer set $\mathcal{P} = \{p_1, p_2, \dots, p_k\}$, where token pruning is performed sequentially. 
For each pruning layer $p_i \in \mathcal{P}$, we compute an accumulated TTV score for each token by summing its TTV values from all preceding accumulation layers up to and including $p_i$:
\begin{equation}
\text{TTV}_{p_i}(T_{I}) = \sum_{l \in \mathcal{A}, l \leq p_i} \text{TTV}_l(T_{I}).
\end{equation}
This accumulation strategy enables each pruning stage to make decisions based on the transition history of each token, leading to more precise pruning.

% Moreover, unlike PDrop~\cite{pdrop}, which follows a pyramid pruning strategy by removing more tokens in early layers and fewer in later ones, we adopt an inverted pyramid strategy.
% This is based on our observation that TTV scores are less reliable in shallow layers.
% To improve pruning reliability, we retain more tokens in early layers and gradually increase the pruning ratio in deeper layers.

\begin{table*}[t]
    \centering
    %\gustavo{some columns do not have bold font numbers.}
    \begin{adjustbox}{max width=\textwidth}
    \begin{tabular}{>{\raggedright\arraybackslash}m{4cm}|>{
    \centering\arraybackslash}m{2.5cm}>{\centering\arraybackslash}m{1.8cm}|>{\centering\arraybackslash}m{1.5cm}>{\centering\arraybackslash}m{1.5cm}>{\centering\arraybackslash}m{1.5cm}>{\centering\arraybackslash}m{1.5cm}>{\centering\arraybackslash}m{1.5cm}>{\centering\arraybackslash}m{1.5cm}>{\centering\arraybackslash}m{1.5cm}>{\centering\arraybackslash}m{1.5cm}}
    \toprule

    \textbf{Methods} &TFLOPs  & Acc.(\%) & MME$^{P}$ & VQA$^{V2}$ & Seed$^I$ & TextVQA & SQA$^I$ & POPE & GQA & MMB$^{en}$  \\
    \midrule
    \multicolumn{11}{l}{\textit{Upper Bound (100\% TFLOPs)}}\\
    \midrule
    LLaVA-1.5-7B & 3.82 (100\%) & 100.0 (-0.0) & 1506 & 78.5 & 66.2 & 58.2  & 69.5 & 85.9 & 61.9 & 64.6  \\
    \midrule
    \multicolumn{11}{l}{\textit{Approximately 40-50\% TFLOPs}} \\
    \midrule
    FastV$_{K=2,R=0.5}$ \texttt{\scriptsize{(ECCV24)}} & 2.01 (52.6\%) & 97.8 (-2.2) & 1474 & 77.0 & 64.0  & 57.2 & 68.5 & 84.0 & 59.4 & 64.2  \\
    TopV \texttt{\scriptsize{(CVPR25)}} & 1.95 (51.0\%) & - & - & - & - & - & \textbf{69.6} & 84.2 & - & 64.3 \\
    PDrop \texttt{\scriptsize{(CVPR25)}} & 1.78 (46.6\%) & 98.8 (-1.2) & \underline{1500} & \underline{77.7} & 64.3  & 57.5  & 69.4 & 84.8 & 60.1 & \underline{64.9}  \\
    ShortV  \texttt{\scriptsize{(ICCV25)}} & 1.68 (44.0\%) & - & 1342 & - & 62.5 & - & - & - & 58.3 & 60.7 \\
    SparseVLM \texttt{\scriptsize{(ICML25)}} & 1.57 (41.1\%) & 98.8 (-1.2) &1484  & 77.6 & \underline{64.9} &  \textbf{58.0} & 67.7 & \textbf{85.7} & \underline{60.6} & 64.7 \\ 
    \rowcolor{green!20}TransPrune-High (Ours) & 1.56 (40.8\%)  & \textbf{100.0 (-0.0)} & \textbf{1540} & \textbf{77.9} & \textbf{65.7}  & \underline{57.8} & \underline{69.5} & \underline{85.0} & \textbf{61.4} & \textbf{66.0} \\
    \midrule
    \multicolumn{11}{l}{\textit{Approximately 25-35\% TFLOPs}} \\ 
    \midrule
    TopV \texttt{\scriptsize{(CVPR25)}} & 1.34 (35.0\%) & - & - & - & - & - & \textbf{69.5} & 85.0 & - & 60.4 \\
    PDrop \texttt{\scriptsize{(CVPR25)}} & 1.28 (33.5\%)& 96.5 (-3.5) & \underline{1468} & 76.1 & 62.4 & \underline{57.2} & \underline{68.8} & 84.2 & 58.0 & 63.0 \\
    SparseVLM \texttt{\scriptsize{(ICML25)}} & 1.28 (33.5\%) & 97.9 (-2.1) & 1441 & \textbf{77.0} & \underline{64.1} & \textbf{57.8} & 68.7 & \textbf{85.3} & \underline{59.5} & \underline{64.1} \\
    FastV$_{K=2,R=0.75}$ \texttt{\scriptsize{(ECCV24)}} & 1.12 (29.3\%) & 94.4 (-5.6) & 1394 & 74.3 & 61.2 & 56.2 & 68.7 & 79.2 & 56.6 & 62.3 \\
    \rowcolor{green!20}TransPrune-Low (Ours) & 1.19 (31.2\%) & \textbf{98.4 (-1.6)}  & \textbf{1491} & \underline{76.6} &  \textbf{64.2} & 56.5 & 68.7 & \underline{85.1} & \textbf{60.0} & \textbf{65.6}  \\
    \bottomrule
    \end{tabular}
    \end{adjustbox}
    \caption{Performance of \textit{\textbf{within-LLM}} methods across different benchmarks on LLaVA-1.5-7B. TransPrune-High and TransPrune-Low achieve the best performance under low TFLOPs settings. Bold font highlights the best-performing results, and underlined values denote the second-best performance.}
    \label{tab:llava1.5} 
\end{table*}

\begin{table*}[t]
    \centering
    \begin{adjustbox}{max width=\textwidth}
    \begin{tabular}{>{\raggedright\arraybackslash}m{4cm}|>{\centering\arraybackslash}m{2.5cm}>{\centering\arraybackslash}m{1.8cm}|>{\centering\arraybackslash}m{1.5cm}>{\centering\arraybackslash}m{1.5cm}>{\centering\arraybackslash}m{1.5cm}>{\centering\arraybackslash}m{1.5cm}>{\centering\arraybackslash}m{1.5cm}>{\centering\arraybackslash}m{1.5cm}>{\centering\arraybackslash}m{1.5cm}>{\centering\arraybackslash}m{1.5cm}}
    \toprule

    \textbf{Methods} & TFLOPS  & Acc.(\%) & MME$^{P}$ & VQA$^{V2}$ & Seed$^I$ & TextVQA & SQA$^I$ & POPE & GQA & MMB$^{en}$  \\
    \midrule
    \multicolumn{11}{l}{\textit{Upper Bound (100\% TFLOPs)}} \\ 
    \midrule
    LLaVA-Next-7B & 20.83 (100\%) & 100.0 (-0.0) & 1520 & 81.8 & 70.2 & 61.3  & 70.2 & 86.5 & 64.3 & 67.9  \\
    \midrule
    \multicolumn{11}{l}{\textit{Approximately 40-50\% TFLOPs}} \\
    \midrule
    ShortV \texttt{\scriptsize{(ICCV25)}} & 10.62 (51.0\%) & - & \underline{1525} & - & \textbf{70.4} & - & - & - & 63.4 & 67.2 \\
    FastV$_{K=2,R=0.5}$ \texttt{\scriptsize{(ECCV24)}}  & 10.55 (50.6\%) & 98.9 (-1.1) & 1524 & 80.7 & 69.1 & 59.3 & \textbf{69.2} & 86.6 & \textbf{63.6} & \textbf{67.8}  \\
    PDrop \texttt{\scriptsize{(CVPR25)}} & 9.46 (45.4\%) & 99.3 (-0.7) & 1511 & \textbf{81.2} & 69.0  & \textbf{61.8}  & 69.0 & \underline{86.7} & 63.3 & 67.4  \\
    \rowcolor{green!20}TransPrune-High (Ours) & 8.33 (40.0\%) & \textbf{99.8 (-0.2)} & \textbf{1528} & \underline{81.1} & \underline{70.1}  & \textbf{61.8} & \underline{69.1} & \textbf{86.9} & \textbf{63.6} & \textbf{67.8}  \\
    \midrule
    \multicolumn{11}{l}{\textit{Approximately 25-35\% TFLOPs}} \\
    \midrule
    PDrop \texttt{\scriptsize{(CVPR25)}} & 6.65 (31.9\%) & 98.2 (-1.8) & \underline{1492} & \textbf{80.2} & \underline{68.4} & \textbf{60.2} & \underline{68.3} & \textbf{86.6} & \textbf{62.7} & \underline{67.2} \\
    FastV$_{K=2,R=0.75}$ \texttt{\scriptsize{(ECCV24)}} & 5.80 (27.8\%) & 95.5 (-4.5) & 1465 & 78.4 & 66.4 & 57.4 & 67.5 & 83.7 & 60.6 & 65.6 \\
    \rowcolor{green!20}TransPrune-Low (Ours) & 6.41 (30.8\%) & \textbf{98.4 (-1.6)} & \textbf{1500} & \underline{80.0} & \textbf{70.2}  & \underline{60.1} & \textbf{68.4} & \textbf{86.6} & \underline{61.5} & \textbf{67.3} \\
    \bottomrule
    \end{tabular} %}
    \end{adjustbox}
    
    \caption{Performance of \textit{\textbf{within-LLM}} methods across different benchmarks on LLaVA-Next-7B. TransPrune-High and TransPrune-Low achieve the best performance under low TFLOPs settings.}
    \label{tab:llava1.6} 
\end{table*}

\begin{table}[t]
    \centering
    \begin{adjustbox}{max width=\linewidth}
    \begin{tabular}{>{\raggedright\arraybackslash}m{2.8cm}|>{\centering\arraybackslash}m{1.5cm}|>{\centering\arraybackslash}m{1.5cm}>{\centering\arraybackslash}m{1.5cm}>{\centering\arraybackslash}m{1.5cm}>{\centering\arraybackslash}m{1.5cm}}
    \toprule

    \textbf{Methods} &TFLOPs & MME$^{P}$ & SQA$^I$ & POPE & MMB$^{en}$ \\
    \midrule
    Qwen2.5-VL-7B & 100\% & 1634 & 79.6 & 86.2 & 79.8 \\
    \midrule
    FastV & 53.6\% & 1563 & 78.3 & 85.1 &77.6 \\
    \rowcolor{green!20}TransPrune (Ours)& 45.1\% & 1580 & 78.1 & 87.5  & 78.1 \\
    \bottomrule
    \end{tabular} 
    \end{adjustbox}
    
    \caption{Performance of \textit{\textbf{within-LLM}} methods across different benchmarks on Qwen2.5-VL-7B.}
    \label{tab:qwen2.5VL} 
\end{table}

\subsection{Instruction-Guided Attention}
Since TTV relies solely on the intrinsic variation of image tokens and is independent of the instruction, it may fail to capture instruction-related information. 
To address this issue, we introduce \textbf{Instruction-Guided Attention (IGA)}.
We simply leverage how instruction tokens attend to image tokens to estimate the importance of each image token.

Specifically, we first compute the attention matrix $A$ between the query of the instruction tokens $Q$ and the key of image tokens $K_{T_{I}}$.
We then average them over all instruction tokens to obtain the IGA:
\begin{equation}
\text{IGA}(T_{I}) = \frac{1}{L} \sum_{j=1}^{L} A_j,
\end{equation}
where $A_{j}$ indicates the weight of attention from the $j$-th instruction token to the image tokens and $L$ indicates the length of instruction tokens.
A higher IGA score indicates that the token is more semantically relevant under the given instruction. 

For each pruning layer $p_i$, TransPrune integrates the accumulated TTV with IGA to determine token importance for pruning. 
The combined pruning score for image tokens $T_{I}$ at layer $p_i$ is computed as:
\begin{equation}
    \text{Score}_{p_i}(T_{I}) = \alpha \cdot \text{TTV}_{p_i}(T_{I}) + (1 - \alpha) \cdot \text{IGA}_{p_{i}+1}(T_{I}),
\end{equation} 
where hyperparameter $\alpha \in [0,1]$ balances the contributions of TTV and IGA. 
Tokens with lower combined scores are subsequently pruned.
Note that the accumulation mechanism is exclusively applied to TTV and is not utilized in IGA.

\section{Experiment}

\begin{table*}[t]\small
    \centering
    \begin{adjustbox}{max width=\textwidth}
    \begin{tabular}{>{\raggedright\arraybackslash}m{3.5cm}|>{\centering\arraybackslash}m{1.8cm}>{\centering\arraybackslash}m{1.8cm}>{\centering\arraybackslash}m{2cm}|>{\centering\arraybackslash}m{1.3cm}>{\centering\arraybackslash}m{1.3cm}>{\centering\arraybackslash}m{1.3cm}>{\centering\arraybackslash}m{1.3cm}>{\centering\arraybackslash}m{1.3cm}>{\centering\arraybackslash}m{1.3cm}}
    \toprule
    \textbf{Methods} & Final Token & TFLOPs & Acc.(\%) & MME$^P$ & SQA$^I$ & GQA & POPE & MMB$^{en}$ & Seed$^I$ \\
    \midrule
    \multicolumn{10}{l}{\textit{Upper Bound (100\% TFLOPs, 576 tokens)}} \\ 
    \midrule
    LLaVA-1.5-7B &576&3.82 (100\%)& 100&1506&69.5&61.9&85.9&64.6&66.2 \\
    \midrule
    \multicolumn{10}{l}{\textit{Retained 36 tokens}} \\
    \midrule
    VisionZip \texttt{\scriptsize{(CVPR25)}} & 288 & 1.89 (49.5\%) & 98.5 & \underline{1457} & \underline{68.8} & \textbf{60.3} & \textbf{86.3} & \textbf{64.3} & \textbf{64.7}  \\
    VisionZip+FastV & 144 & 1.00 (26.2\%)& 95.8 (-2.7) &1423 &68.7&58.0&83.0&62.5&62.5\\
    VisionZip+PDrop & 36 & 0.88 (23.0\%) & 96.4 (-2.1)& 1447 & 68.7 & 58.2 & 84.5 & 63.5 & 61.7     \\
    \rowcolor{green!20}VisionZip+TransPrune  & 36 & 0.66 (17.3\%) & \textbf{98.0 (-0.5)} & \textbf{1460} & \textbf{68.9} & \underline{59.4} & \textbf{86.3} & \underline{63.8} & \underline{64.1} \\
    \midrule
    \multicolumn{10}{l}{\textit{Retained 24 tokens}} \\
    \midrule
    VisionZip \texttt{\scriptsize{(CVPR25)}} & 192 & 1.25 (32.7\%) & 97.2 & \underline{1443} & 68.8 & \underline{59.3} & \underline{85.5} & \underline{62.9} & \textbf{63.2}  \\
    VisionZip+FastV & 96 & 0.66 (17.3\%) & 94.3 (-2.9) & 1383 & \underline{69.0} & 56.8 & 81.1& 62.3 & 61.0 \\
    VisionZip+PDrop & 24 & 0.59 (15.4\%) & 94.8 (-2.4)& 1417 & \textbf{69.7} & 56.8 & 83.2 & 61.2& 60.3  \\
    \rowcolor{green!20}VisionZip+TransPrune  & 24 & 0.44 (11.5\%) & \textbf{97.2 (-0.0)} & \textbf{1444} & 68.6 & \textbf{59.4} & \textbf{85.9} & \textbf{63.0} & \underline{62.7} \\
    \bottomrule
    \end{tabular}
    \end{adjustbox}
    
    \caption{Performance when combined with the \textit{\textbf{projector-based}} method VisionZip.
    Our method achieves a reduction in FLOPs while maintaining performance comparable to VisionZip alone.}
    \label{tab:visionzip} 
\end{table*}

\begin{table*}[t]\small
    \centering
    \begin{adjustbox}{max width=\textwidth}
    \begin{tabular}{>{\raggedright\arraybackslash}m{3.5cm}|>{\centering\arraybackslash}m{1.8cm}>{\centering\arraybackslash}m{1.8cm}>{\centering\arraybackslash}m{2cm}|>{\centering\arraybackslash}m{1.3cm}>{\centering\arraybackslash}m{1.3cm}>{\centering\arraybackslash}m{1.3cm}>{\centering\arraybackslash}m{1.3cm}>{\centering\arraybackslash}m{1.3cm}>{\centering\arraybackslash}m{1.3cm}}
    \toprule
    \textbf{Methods} & Final Token & TFLOPs & Acc.(\%) & MME$^P$ & SQA$^I$ & GQA & POPE & MMB$^{en}$ & Seed$^I$ \\
    \midrule
    \multicolumn{10}{l}{\textit{Upper Bound (100\% TFLOPs, 576 tokens)}} \\ 
    \midrule
    LLaVA-1.5-7B &576&3.82 (100\%)& 100&1506&69.5&61.9&85.9&64.6&66.2 \\
    \midrule
    \multicolumn{10}{l}{\textit{Retained 36 tokens}} \\
    \midrule
    CDPruner \texttt{\scriptsize{(NeurIPS25)}} & 288 & 1.89 (49.5\%) & 98.8 & 1452 & \textbf{68.6} & \textbf{60.8} & \textbf{86.9} & \textbf{64.2} &  \textbf{65.6} \\
    CDPruner+FastV & 144 & 1.00 (26.2\%) & 96.3 (-2.5) & 1440 & 67.5 & 58.7 & 84.3 & 62.4 & 63.4 \\
    CDPruner+PDrop & 36 & 0.88 (23.0\%) & 96.7 (-2.1) & \underline{1455} & \underline{68.0} & 59.0 & 85.2 & 62.4 & 62.5 \\
    \rowcolor{green!20}CDPruner+TransPrune  & 36 & 0.66 (17.3\%) & \textbf{98.3 (-0.5)} & \textbf{1467} & 67.9 & \underline{60.4} & \underline{86.8} & \underline{63.8} & \underline{64.6}  \\
    \midrule
    \multicolumn{10}{l}{\textit{Retained 24 tokens}} \\
    \midrule
    CDPruner \texttt{\scriptsize{(NeurIPS25)}} & 192 & 1.25 (32.7\%) & 98.3 & \textbf{1447} & \underline{68.8} & \textbf{60.3} & \textbf{87.3} & \textbf{63.1} & \textbf{64.7} \\
    CDPruner+FastV & 96 & 0.66 (17.3\%) & 96.2 (-2.1) & 1419 & \underline{68.8} & 58.2 & 84.9 & 62.1 & 62.9  \\
    CDPruner+PDrop & 24 & 0.59 (15.4\%) & 95.9 (-2.4) & 1407 & \textbf{69.2} & 57.9 & 85.5 & 62.3 &  61.6 \\
    \rowcolor{green!20}CDPruner+TransPrune  & 24 & 0.44 (11.5\%) & \textbf{97.6 (-0.7)} & \underline{1430} & 68.7 & \underline{59.3} & \textbf{87.3} & \textbf{63.1} & \underline{64.1}  \\
    \bottomrule
    \end{tabular}
    \end{adjustbox}
    
    \caption{Performance when combined with the \textit{
    \textbf{projector-based}} method CDPruner.
    Our method achieves a reduction in FLOPs while maintaining performance comparable to CDPruner alone.}
    \label{tab:cdpruner} 
\end{table*}

\begin{table}[ht]
    \centering
    \begin{adjustbox}{max width=\linewidth}
    \begin{tabular}{
        >{\raggedright\arraybackslash}m{3.5cm}|
        >{\centering\arraybackslash}m{1.3cm}|
        >{\centering\arraybackslash}m{1cm}
        >{\centering\arraybackslash}m{1cm}|
        >{\centering\arraybackslash}m{1cm}
        >{\centering\arraybackslash}m{1cm}
        }
    \toprule
    \multirow{2}{*}{Methods} & \multirow{2}{*}{TFLOPs} & \multicolumn{2}{c|}{TGIF} & \multicolumn{2}{c}{MSVD} \\
    \cmidrule(lr){3-4} \cmidrule(lr){5-6}
     &  & Acc & Score & Acc & Score \\
    \midrule
    Video-LLaVA & 14.4 & 47.0 & 3.3 & 69.6 & 3.9 \\
    \midrule
    w / FastV & 7.4 & 47.6 & 3.4 & 70.3 & 3.9 \\
    w / PDrop & 6.6 & 46.9 & 3.4 & 70.0 & 3.9 \\
    \rowcolor{green!20}w / TransPrune (Ours) & 6.1 & 47.5 & 3.4 & 70.3 & 3.9 \\
    \bottomrule
    \end{tabular}
    \end{adjustbox}
    
    \caption{Performance of \textit{\textbf{within-LLM}} methods across different benchmarks on Video-LLaVA.} 
    \label{tab:video}
\end{table}

\begin{table}[ht]\small
    \centering
    \begin{adjustbox}{max width=\linewidth}
    \begin{tabular}{>{\raggedright\arraybackslash}m{2cm}|>{\centering\arraybackslash}m{1.5cm}>{\centering\arraybackslash}m{1.5cm}>{\centering\arraybackslash}m{1.5cm}>{\centering\arraybackslash}m{1.5cm}>{\centering\arraybackslash}m{1.5cm}}
    \toprule
    \textbf{Methods} & SQA$^I$ & GQA & POPE & MMB$^{en}$ & TextVQA \\
    \midrule
    LLaVA-1.5-7B & 69.5 & 61.9 & 85.9 & 64.6 & 58.2 \\
    TTV-only & 68.9 & 58.4 & 82.1 & 64.9 & 50.9 \\
    \bottomrule
    \end{tabular}
    \end{adjustbox}
   
    \caption{Performance using only TTV.}
    \label{tab:onlyttv} 
\end{table}

\subsection{Setup}
\noindent{\textbf{Benchmarks.}}
To thoroughly evaluate the effectiveness of TransPrune, we conduct experiments across a diverse set of benchmarks spanning tasks such as perception, reasoning, and visual question answering (VQA).
The benchmarks include:
MME~\cite{mme}, 
MMBench~\cite{mmbench}, 
SEED~\cite{seed}, 
ScienceQA~\cite{scienceqa}, 
VQA-v2~\cite{goyal2017making}, 
POPE~\cite{gqa},
GQA~\cite{gqa}
and TextVQA \citep{singh2019towards}.
In addition, we select two video benchmarks, TGIF and MSVD, to evaluate the generalization capability of TransPrune in Video LLM.

\noindent{\textbf{Models.}}
We validate the effectiveness and generalization capability of TransPrune through extensive experiments on LVLMs with diverse architectural designs and input resolutions.
Specifically, our study includes LLaVA-v1.5-7B~\cite{liu2024visual}, LLaVA-NeXT-7B~\cite{liu2024llavanext} and Qwen2.5-VL-7B~\cite{qwen2.5vl}. 
For video, we select Video-LLaVA~\cite{video-llava} as base model.

\noindent{\textbf{Methods.}}
As TransPrune is a within-LLM method, we conduct a fair comparison with existing within-LLM pruning approaches, including FastV~\cite{fastv}, TopV~\cite{topv}, PDrop~\cite{pdrop}, ShortV~\cite{shortv}, and SparseVLM~\cite{sparsevlm}. 
We further demonstrate its potential when combined with two representative projector-based pruning methods, VisionZip~\cite{visionzip} and CDPruner~\cite{cdpruner}.

\noindent{\textbf{Implementation details.}}
We set $\alpha$=0.5 to equally balance the contributions of TTV and IGA. 
TTV is accumulated across layers 7 to 12, while token pruning is performed at layers 7, 9, and 12. 
To evaluate the effectiveness of TransPrune under different computational budgets, we design two configurations—TransPrune-High and TransPrune-Low—which keep different numbers of tokens at each pruning layer (see the supplementary material).
All experiments are conducted on A100 GPUs (40GB). 
During inference, we leverage \textbf{FlashAttention}~\cite{flashattention} for efficient attention computation.
Since TransPrune’s TTV computation only requires access to module inputs and outputs, and IGA exclusively computes attention weights from instruction tokens to image tokens (rather than full attention maps), our method remains compatible with FlashAttention.

\subsection{Comparison with SOTA in Public Benchmarks}
TransPrune achieves strong performance across a wide range of benchmarks while incurring low TFLOPs among all compared methods.
As shown in Table~\ref{tab:llava1.5}, TransPrune-High maintains negligible performance degradation while reducing computational cost to just 41\% of the original TFLOPs.
Furthermore, as shown in Table~\ref{tab:llava1.6} on the higher-resolution LLaVA-NeXT-7B, TransPrune maintains even lower TFLOPs while simultaneously outperforming other methods.
For Qwen2.5-VL, which has an architecture different from LLaVA, our method also demonstrates strong performance, validating the generalization capability of TransPrune.

While TransPrune is a within-LLM pruning method, it also integrates seamlessly with existing projector-based token pruning approaches.
As shown in the Table~\ref{tab:visionzip} and Table~\ref{tab:cdpruner}, we present its combined performance with VisionZiP~\cite{visionzip} and CDPruner~\cite{cdpruner}. 
TransPrune achieves consistently strong results, demonstrating its compatibility and effectiveness with existing projector-based pruning methods.
When retaining only 24 tokens, the combination with VisionZiP~\cite{visionzip} achieves a substantial reduction in TFLOPs with almost no performance degradation.

\subsection{Analysis}
We use TransPrune-High for the following analysis.

\noindent{\textbf{Effectiveness of TTV.}}
To assess the effectiveness of TTV as a standalone criterion for token importance, we conduct experiments using only TTV for token pruning.
As shown in Table~\ref{tab:onlyttv}, our results demonstrate that the attention-independent TTV also achieves competitive performance, highlighting its effectiveness as a standalone criterion for token importance.
However, we also observe a noticeable performance drop when using TTV alone on TextVQA. 
This may be attributed to TTV’s exclusive reliance on image token transitions, as it does not account for instruction-dependent semantics~\cite{problem}.

\begin{figure}[!ht]
    \centering
    \includegraphics[width=1.0\linewidth]{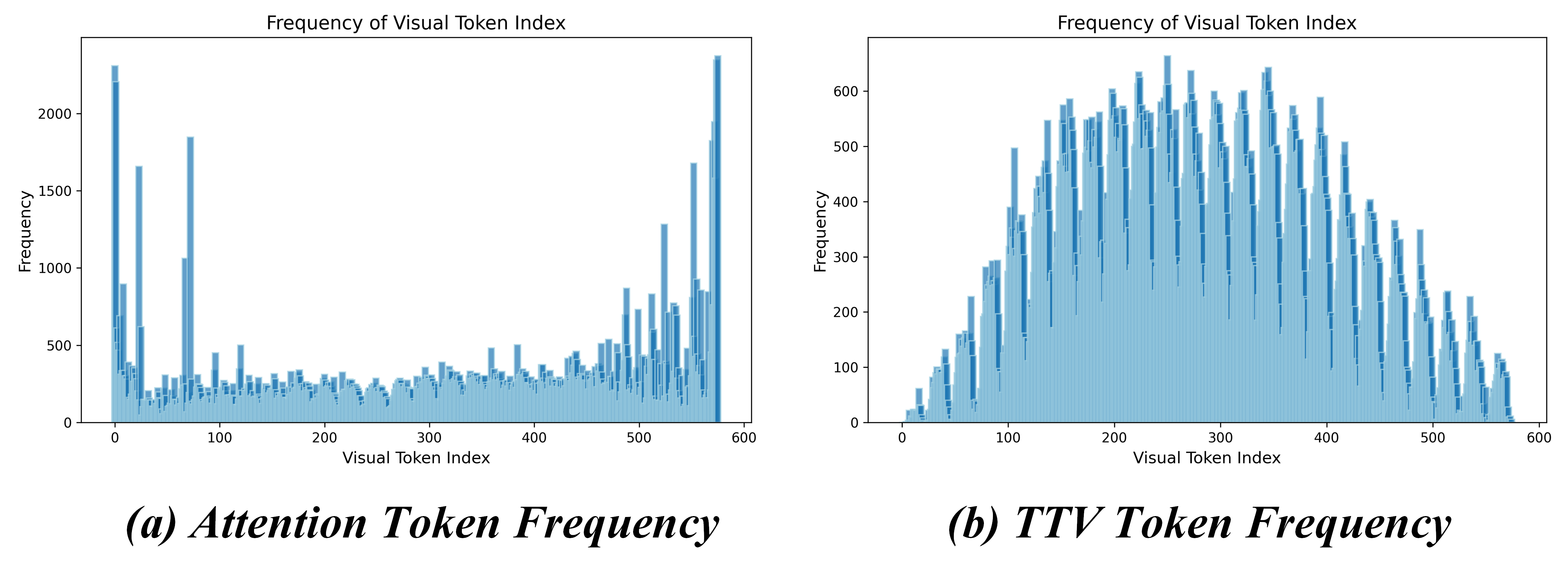}
    \caption{
    Token position frequency statistics on MME benchmark for IGA and TTV. 
    } 
    \label{fig:position}
\end{figure}

Besides, we visualize the positional distribution of the final retained tokens across all MME samples, as shown in Figure~\ref{fig:position}. 
Figure~\ref{fig:position}~(a) presents the frequency using IGA. 
Since IGA is an attention-based method, it clearly exhibits positional bias, favoring the retention of tokens at the beginning and end~\cite{problem,pact}. 
However, for images, tokens at these positions often carry less semantic information. 
In contrast, Figure~\ref{fig:position}~(b) shows the frequency using TTV. 
TTV introduces no apparent positional bias and tends to focus more uniformly on the central regions of the image, which typically encapsulate denser and more relevant semantic information.
The combined use of TTV and attention-based methods can partially alleviate the issue of positional bias.

\noindent{\textbf{Efficiency of TransPrune.}}
We analyze the additional computational cost introduced by TransPrune in a setting where there are $ l $ instruction tokens, each with hidden dimension $ d $, and $ m $ denotes the intermediate dimension of the FFN layer.
In common VQA tasks, the instruction typically consists of only a few dozen tokens.
TransPrune operates in $ s $ \emph{pruning stages}, where at each stage, a subset of $ n_i $ visual tokens is retained (with $ i = 1, 2, ..., s $). 
Each pruning stage may correspond to multiple Transformer layers. 
Let $ k_i $ denote the number of layers in stage $ i $.
The extra computations introduced by TransPrune mainly come from two components: (1) the L2 norm and cosine similarity calculations in TTV, and (2) the attention between retained visual tokens and instruction tokens in IGA.
Formally, the total FLOPs can be approximated as:
\begin{equation}
\sum_{i=1}^{s} k_i \left( 4 n_i d^2 + 2 n_i^2 d + 3 n_i d m \right) + \sum_{i=1}^{s-1} l n_i d + \mathcal{O}(s d),
\end{equation}
where the first term corresponds to the Transformer operations  on the retained tokens across all layers, the second term represents the attention between instruction tokens and visual tokens at each stage, and the last term captures the small overhead from TTV computations, which scales linearly with the number of stages and token dimension.
Compared to the baseline model’s total computation, the extra cost introduced by TransPrune is marginal.
Besides, we evaluate the latency (ms) and memory usage (GB) of various methods on the MME benchmark under identical experimental settings. TransPrune demonstrates superior performance with reduced latency and memory consumption.

\begin{table}[ht]
    \centering
    \footnotesize
    \begin{adjustbox}{max width=\linewidth}
    \begin{tabular}{>{\raggedright\arraybackslash}m{1.5cm}|>{\centering\arraybackslash}m{1.5cm}>{\centering\arraybackslash}m{1.8cm}>{\centering\arraybackslash}m{1cm}}
    \toprule
    Methods & Latency (ms) & Memory (GB) & Accuracy \\
    \midrule
    FastV       & 125.2 & 14.99 & 1474 \\
    PDrop       & 115.2 & 14.87 & 1500 \\
    SparseVLM   & 129.1 & 19.05 & 1484 \\
    \rowcolor{green!20}TransPrune        & \textbf{111.4} & \textbf{14.82} & \textbf{1540} \\
    \bottomrule
    \end{tabular}
    \end{adjustbox}
    \caption{Comparison of latency, memory, and accuracy for different pruning methods in MME.}
    \label{tab:efficiency} 
\end{table}

\begin{figure}[!ht]
    \centering
    \includegraphics[width=1.0\linewidth]{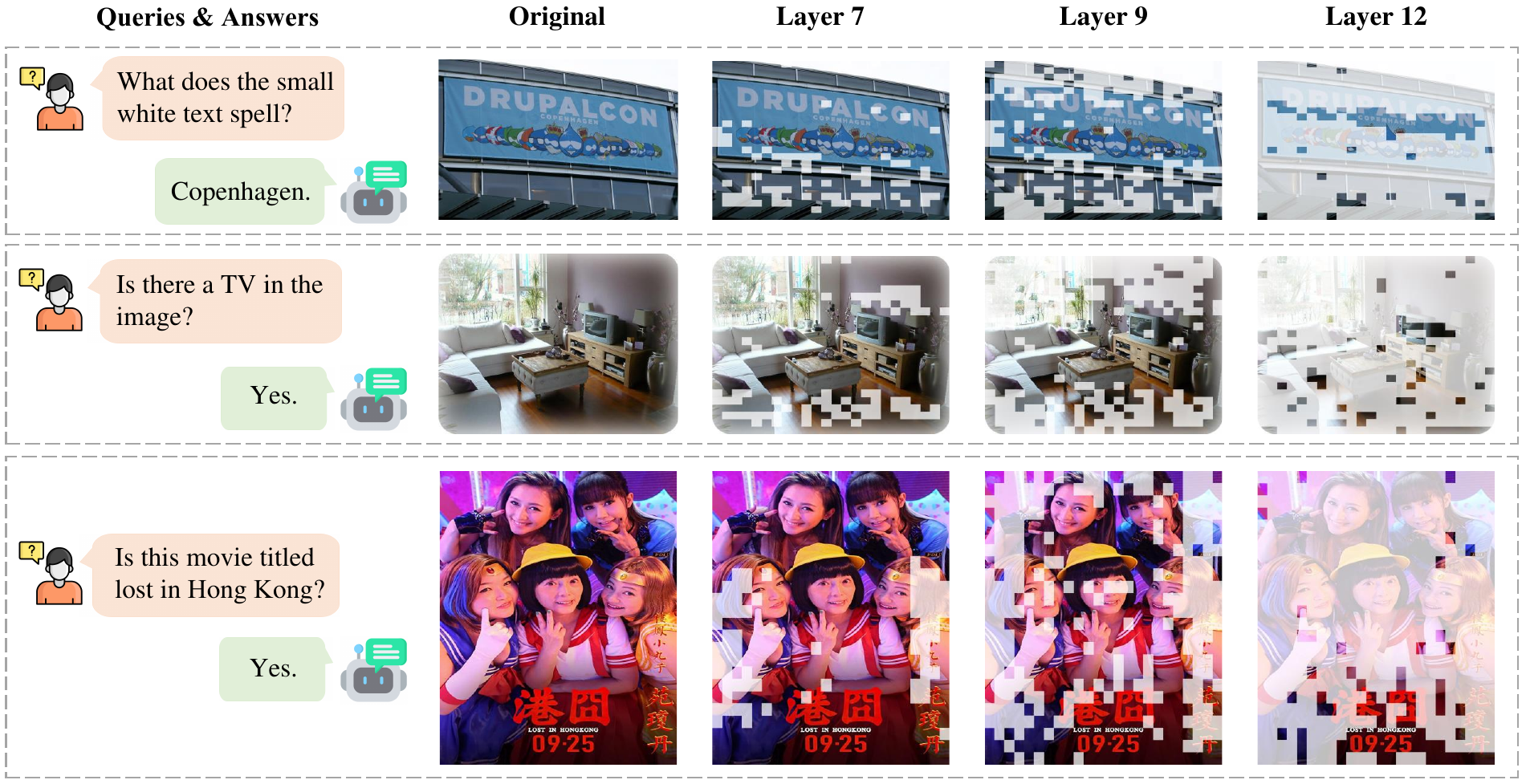}
    \caption{Visualization of TransPrune on different VQA prompts.} 
    \label{fig:case}
\end{figure}

\noindent{\textbf{Qualitative visualization.}}
As shown in Figure~\ref{fig:case}, we present token pruning results across different layers for three VQA examples. 
From left to right, we visualize how the retained tokens evolve as pruning progresses.

In all examples, less relevant tokens are progressively discarded, while semantically important tokens are consistently preserved at the final pruning stage.

\noindent\textbf{Impact of different layer choices.}
Since the accumulation layers and pruning layers are two critical hyperparameters in TransPrune, we conduct experiments to evaluate their impact on performance.
The pruning layers are determined by the accumulation layers, making it difficult to analyze their effects independently. 
Therefore, we consider them jointly.
As shown in Table~\ref{tab:pruning_layers}, the combination of layers 7, 9, and 12 achieves the best performance.
This result corroborates our previous analysis, confirming that transitions in the middle layers provide a more reliable signal of token semantic importance.

\begin{table}[ht]
    \centering
    \small
    \begin{adjustbox}{max width=\linewidth}
    \begin{tabular}{>{\centering\arraybackslash}m{1.2cm}|>{\centering\arraybackslash}m{1.2cm}>{\centering\arraybackslash}m{1.2cm}>{\centering\arraybackslash}m{1.2cm}|>{\centering\arraybackslash}m{1.5cm}>{\centering\arraybackslash}m{1.5cm}}
    \toprule
    \multicolumn{1}{c|}{Group} & Layer 1 & Layer 2 & Layer 3 & MME & TFLOPs \\
    \midrule
    1 & 5  & 7  & 10 & 1479 & 1.35 \\
    2 & 6  & 8  & 11 & 1496 & 1.45 \\
    \rowcolor{green!20}3 & 7  & 9  & 12 & \textbf{1540} & \textbf{1.56} \\
    4 & 8  & 10 & 13 & 1493 & 1.66 \\
    5 & 9  & 11 & 14 & 1497 & 1.77 \\
    \bottomrule
    \end{tabular}
    \end{adjustbox}
    \caption{Comparison of MME performance and TFLOPs under different layer selections. 
    For example, the entry ‘5, 7, 10’ indicates that pruning is performed at layers 5, 7, and 10, with the corresponding accumulation layers spanning from 5 to 10.}
    \label{tab:pruning_layers} 
\end{table}

\noindent\textbf{Shallow vs. deep accumulation layers on TTV.}
To investigate the effect of accumulation layers on TTV, we compare token pruning using TTV computed from deeper layers (layers 7–12) with that computed solely from shallow layers (layers 1–6).
In all settings, token pruning is performed at layers 7, 9, and 12 to ensure a fair comparison.
For the shallow-layer setting, TTV at each pruning layer is accumulated only from the preceding shallow layers: for example, at pruning layer 7, we use TTV from layer 1; at layer 9, from layers 1–3; and at layer 12, from layers 1–6.
As reported in Table~\ref{tab:layerc}, TTV computed solely from shallow layers is less effective than TTV derived from deeper layers.
These results indicate that tokens with larger transitions in the middle layers are more semantically informative than those in shallow layers, because the middle layers combine shallow global information with deep local details, thereby more effectively reflecting the LLM’s shifting attention.
\begin{table}[ht]
    \centering
    \footnotesize
    \begin{adjustbox}{max width=\linewidth}
    \begin{tabular}{>{\raggedright\arraybackslash}m{2cm}|>{\centering\arraybackslash}m{1cm}>{\centering\arraybackslash}m{1cm}>{\centering\arraybackslash}m{1cm}>{\centering\arraybackslash}m{1cm}}
    \toprule
    Layers  & MME$^P$ & SQA$^I$ & GQA & MMB$^{en}$  \\
    \midrule
    Layers (1-6) & 1515 & 69.4 & 61.3 & 65.6 \\
    \rowcolor{green!20}Layers (7-12) & \textbf{1540} & \textbf{69.5} & \textbf{61.4} & \textbf{66.0} \\
    \bottomrule
    \end{tabular} %}
    \end{adjustbox}
    \caption{Impact of accumulated TTV across different layers.}
    \label{tab:layerc} 
\end{table}

\noindent{\textbf{Impact of accumulation mechanism.}}
To verify the effectiveness of the accumulation mechanism, we conduct ablation experiments using the same pruning layers as TransPrune, as shown in Table~\ref{tab:accumulation}.
Almost all benchmarks show improvements after introducing the accumulation mechanism, indicating that it enables TTV to capture more precise semantic information.
\begin{table}[ht]
    \centering
    \footnotesize
    \begin{adjustbox}{max width=\linewidth}
    \begin{tabular}{>{\raggedright\arraybackslash}m{2.5cm}|>{\centering\arraybackslash}m{1cm}>{\centering\arraybackslash}m{1cm}>{\centering\arraybackslash}m{1cm}>{\centering\arraybackslash}m{1cm}}
    \toprule
    Methods  & MME$^P$ & SQA$^I$ & GQA & MMB$^{en}$  \\
    \midrule
    w/o Accumulation & 1530 & 69.2 & \textbf{61.4} & 65.7 \\
    \rowcolor{green!20}w Accumulation & \textbf{1540}  & \textbf{69.5}  & \textbf{61.4} & \textbf{66.0} \\
    \bottomrule
    \end{tabular} %}
    \end{adjustbox}
    \caption{Ablation study on the impact of accumulation.}
    \label{tab:accumulation} 
\end{table}

\noindent{\textbf{Impact of magnitude and direction.}}
To evaluate the contribution of the magnitude and direction components within TTV, we perform an ablation study by progressively incorporating these elements, starting from a baseline using only IGA.
As presented in the Table~\ref{tab:asdm}, both magnitude and direction contribute to performance gains, with magnitude yielding a more significant improvement. 
Combining both components leads to the optimal performance.
\begin{table}[ht]
    \centering
    \footnotesize
    \begin{adjustbox}{max width=\linewidth}
    \begin{tabular}{>{\raggedright\arraybackslash}m{2cm}|>{\centering\arraybackslash}m{1cm}>{\centering\arraybackslash}m{1cm}>{\centering\arraybackslash}m{1cm}>{\centering\arraybackslash}m{1cm}}
    \toprule
    Methods  & MME$^P$ & SQA$^I$ & GQA & MMB$^{en}$  \\
    \midrule
    Only IGA & 1514 & 69.0 & 61.1 & 65.6 \\
    IGA+Direction  & 1521 & 69.1 & 61.2 & 65.4  \\
    IGA+Magnitude & 1532 & 69.4 & \textbf{61.4} & 65.7 \\
    \rowcolor{green!20}IGA+TTV  & \textbf{1540} & \textbf{69.5} & \textbf{61.4} & \textbf{66.0} \\
    \bottomrule
    \end{tabular} %}
    \end{adjustbox}
    \caption{Ablation study on the impact of direction and magnitude.}
    \label{tab:asdm} 
\end{table}

\noindent{\textbf{Impact of the parameter $\alpha$.}}
We conduct experiments on the impact of different $\alpha$ parameters on the final performance, as shown in Table~\ref{tab:pa}.
When $\alpha$ is set to 0.5, meaning TTV and IGA contribute equally, the performance reaches its optimal level.
This shows that balancing the contributions of TTV and IGA allows the model to fully leverage both the image token’s own information and the instruction information, leading to the best overall performance.
\begin{table}[ht]
    \centering
    \footnotesize
    \begin{adjustbox}{max width=\linewidth}
    \begin{tabular}{>{\raggedright\arraybackslash}m{1.2cm}|>{\centering\arraybackslash}m{1cm}>{\centering\arraybackslash}m{1cm}>{\centering\arraybackslash}m{1cm}>{\centering\arraybackslash}m{1cm}}
    \toprule
    Parameter & MME$^P$ & SQA$^I$ & GQA & MMB$^{en}$  \\
    \midrule
    $\alpha$=0.4 & \textbf{1540} & 69.4 & \textbf{61.4} & 65.5 \\
    \rowcolor{green!20}$\alpha$=0.5 & \textbf{1540} & \textbf{69.5} & \textbf{61.4} & \textbf{66.0} \\
    $\alpha$=0.6 & 1525 & \textbf{69.5} & \textbf{61.4} & 65.9 \\
    \bottomrule
    \end{tabular}
    \end{adjustbox}
     
    \caption{Ablation study on the impact of parameter $\alpha$.}
    \label{tab:pa}
\end{table}

\section{Conclusion}
In this paper, we explore a novel perspective for LVLM token pruning that is distinct from attention- or similarity-based approaches: leveraging the transition of token representations to reflect token importance.
Based on this insight, we propose TransPrune, a training-free and efficient pruning method.
TransPrune's core relies on combining token transition variation with instruction-guided attention.
Extensive experiments validate the effectiveness and efficiency of TransPrune across a wide range of benchmarks and ablation studies further confirm our findings regarding the importance of middle-layer token transitions.
We believe that this work opens new avenues for accelerating LVLM inference.

{
   \small
   \bibliographystyle{ieeenat_fullname}
   \bibliography{main}

@article{dynamicvit,
  title={Dynamicvit: Efficient vision transformers with dynamic token sparsification},
  author={Rao, Yongming and Zhao, Wenliang and Liu, Benlin and Lu, Jiwen and Zhou, Jie and Hsieh, Cho-Jui},
  journal={Advances in neural information processing systems},
  volume={34},
  pages={13937--13949},
  year={2021}
}

@article{attention,
  title={Attention mechanisms in computer vision: A survey},
  author={Guo, Meng-Hao and Xu, Tian-Xing and Liu, Jiang-Jiang and Liu, Zheng-Ning and Jiang, Peng-Tao and Mu, Tai-Jiang and Zhang, Song-Hai and Martin, Ralph R and Cheng, Ming-Ming and Hu, Shi-Min},
  journal={Computational visual media},
  volume={8},
  number={3},
  pages={331--368},
  year={2022},
  publisher={TUP}
}

@inproceedings{topv,
  title={Topv: Compatible token pruning with inference time optimization for fast and low-memory multimodal vision language model},
  author={Yang, Cheng and Sui, Yang and Xiao, Jinqi and Huang, Lingyi and Gong, Yu and Li, Chendi and Yan, Jinghua and Bai, Yu and Sadayappan, Ponnuswamy and Hu, Xia and others},
  booktitle={Proceedings of the Computer Vision and Pattern Recognition Conference},
  pages={19803--19813},
  year={2025}
}

@inproceedings{transf,
  title={Token transformation matters: Towards faithful post-hoc explanation for vision transformer},
  author={Wu, Junyi and Duan, Bin and Kang, Weitai and Tang, Hao and Yan, Yan},
  booktitle={Proceedings of the IEEE/CVF Conference on Computer Vision and Pattern Recognition},
  pages={10926--10935},
  year={2024}
}

@inproceedings{gqa,
  title={Gqa: A new dataset for real-world visual reasoning and compositional question answering},
  author={Hudson, Drew A and Manning, Christopher D},
  booktitle={Proceedings of the IEEE/CVF conference on computer vision and pattern recognition},
  pages={6700--6709},
  year={2019}
}

@inproceedings{scienceqa,
    title={Learn to Explain: Multimodal Reasoning via Thought Chains for Science Question Answering},
    author={Lu, Pan and Mishra, Swaroop and Xia, Tony and Qiu, Liang and Chang, Kai-Wei and Zhu, Song-Chun and Tafjord, Oyvind and Clark, Peter and Ashwin Kalyan},
    booktitle={The 36th Conference on Neural Information Processing Systems (NeurIPS)},
    year={2022}
}

@article{flashattention,
  title={Flashattention: Fast and memory-efficient exact attention with io-awareness},
  author={Dao, Tri and Fu, Dan and Ermon, Stefano and Rudra, Atri and R{\'e}, Christopher},
  journal={Advances in neural information processing systems},
  volume={35},
  pages={16344--16359},
  year={2022}
}

@inproceedings{visionzip,
  title={Visionzip: Longer is better but not necessary in vision language models},
  author={Yang, Senqiao and Chen, Yukang and Tian, Zhuotao and Wang, Chengyao and Li, Jingyao and Yu, Bei and Jia, Jiaya},
  booktitle={Proceedings of the Computer Vision and Pattern Recognition Conference},
  pages={19792--19802},
  year={2025}
}

@article{problem,
  title={Token Pruning in Multimodal Large Language Models: Are We Solving the Right Problem?},
  author={Wen, Zichen and Gao, Yifeng and Li, Weijia and He, Conghui and Zhang, Linfeng},
  journal={arXiv preprint arXiv:2502.11501},
  year={2025}
}

@misc{pact,
      title={PACT: Pruning and Clustering-Based Token Reduction for Faster Visual Language Models}, 
      author={Mohamed Dhouib and Davide Buscaldi and Sonia Vanier and Aymen Shabou},
      year={2025},
      eprint={2504.08966},
      archivePrefix={arXiv},
      primaryClass={cs.CV},
      url={https://arxiv.org/abs/2504.08966}, 
}

@misc{fastv,
      title={An Image is Worth 1/2 Tokens After Layer 2: Plug-and-Play Inference Acceleration for Large Vision-Language Models}, 
      author={Liang Chen and Haozhe Zhao and Tianyu Liu and Shuai Bai and Junyang Lin and Chang Zhou and Baobao Chang},
      year={2024},
      eprint={2403.06764},
      archivePrefix={arXiv},
      primaryClass={cs.CV},
      url={https://arxiv.org/abs/2403.06764}, 
}

@article{sparsevlm,
  title={Sparsevlm: Visual token sparsification for efficient vision-language model inference},
  author={Zhang, Yuan and Fan, Chun-Kai and Ma, Junpeng and Zheng, Wenzhao and Huang, Tao and Cheng, Kuan and Gudovskiy, Denis and Okuno, Tomoyuki and Nakata, Yohei and Keutzer, Kurt and others},
  journal={arXiv preprint arXiv:2410.04417},
  year={2024}
}

@misc{cdpruner,
      title={Beyond Attention or Similarity: Maximizing Conditional Diversity for Token Pruning in MLLMs}, 
      author={Qizhe Zhang and Mengzhen Liu and Lichen Li and Ming Lu and Yuan Zhang and Junwen Pan and Qi She and Shanghang Zhang},
      year={2025},
      eprint={2506.10967},
      archivePrefix={arXiv},
      primaryClass={cs.CV},
      url={https://arxiv.org/abs/2506.10967}, 
}

@article{vispruner,
      title={Beyond Text-Visual Attention: Exploiting Visual Cues for Effective Token Pruning in VLMs}, 
      author={Zhang, Qizhe and Cheng, Aosong and Lu, Ming and Zhang, Renrui and Zhuo, Zhiyong and Cao, Jiajun and Guo, Shaobo and She, Qi and Zhang, Shanghang},
      journal={arXiv preprint arXiv:2412.01818},
      year={2025},
}

@misc{DivPrune,
      title={DivPrune: Diversity-based Visual Token Pruning for Large Multimodal Models}, 
      author={Saeed Ranjbar Alvar and Gursimran Singh and Mohammad Akbari and Yong Zhang},
      year={2025},
      eprint={2503.02175},
      archivePrefix={arXiv},
      primaryClass={cs.CV},
      url={https://arxiv.org/abs/2503.02175}, 
}

@misc{pdrop,
      title={PyramidDrop: Accelerating Your Large Vision-Language Models via Pyramid Visual Redundancy Reduction}, 
      author={Long Xing and Qidong Huang and Xiaoyi Dong and Jiajie Lu and Pan Zhang and Yuhang Zang and Yuhang Cao and Conghui He and Jiaqi Wang and Feng Wu and Dahua Lin},
      year={2025},
      eprint={2410.17247},
      archivePrefix={arXiv},
      primaryClass={cs.CV},
      url={https://arxiv.org/abs/2410.17247}, 
}

@misc{gridprune,
      title={GridPrune: From "Where to Look" to "What to Select" in Visual Token Pruning for MLLMs}, 
      author={Yuxiang Duan and Ao Li and Yingqin Li and Luyu Li and Pengwei Wang},
      year={2025},
      eprint={2511.10081},
      archivePrefix={arXiv},
      primaryClass={cs.CV},
      url={https://arxiv.org/abs/2511.10081}, 
}

@article{m3d,
  title={M3d: Advancing 3d medical image analysis with multi-modal large language models},
  author={Bai, Fan and Du, Yuxin and Huang, Tiejun and Meng, Max Q-H and Zhao, Bo},
  journal={arXiv preprint arXiv:2404.00578},
  year={2024}
}

@inproceedings{zhang2025individuals,
  title={From Individuals to Crowds: Dual-Level Public Response Prediction in Social Media},
  author={Zhang, Jinghui and Wan, Kaiyang and Xu, Longwei and Li, Ao and Liu, Zongfang and Chen, Xiuying},
  booktitle={Proceedings of the 33rd ACM International Conference on Multimedia},
  pages={5903--5912},
  year={2025}
}

@misc{qwen2.5vl,
      title={Qwen2.5-VL Technical Report}, 
      author={Shuai Bai and Keqin Chen and Xuejing Liu and Jialin Wang and Wenbin Ge and Sibo Song and Kai Dang and Peng Wang and Shijie Wang and Jun Tang and Humen Zhong and Yuanzhi Zhu and Mingkun Yang and Zhaohai Li and Jianqiang Wan and Pengfei Wang and Wei Ding and Zheren Fu and Yiheng Xu and Jiabo Ye and Xi Zhang and Tianbao Xie and Zesen Cheng and Hang Zhang and Zhibo Yang and Haiyang Xu and Junyang Lin},
      year={2025},
      eprint={2502.13923},
      archivePrefix={arXiv},
      primaryClass={cs.CV},
      url={https://arxiv.org/abs/2502.13923}, 
}

@article{emoverse,
  title={EmoVerse: Enhancing Multimodal Large Language Models for Affective Computing via Multitask Learning},
  author={Li, Ao and Xu, Longwei and Ling, Chen and Zhang, Jinghui and Wang, Pengwei},
  journal={Neurocomputing},
  volume={650},
  pages={130810},
  year={2025},
  publisher={Elsevier}
}

@misc{video-xl-pro,
      title={Video-XL-Pro: Reconstructive Token Compression for Extremely Long Video Understanding}, 
      author={Xiangrui Liu and Yan Shu and Zheng Liu and Ao Li and Yang Tian and Bo Zhao},
      year={2025},
      eprint={2503.18478},
      archivePrefix={arXiv},
      primaryClass={cs.CV},
      url={https://arxiv.org/abs/2503.18478}, 
}

@article{mme,
  title={MME: A Comprehensive Evaluation Benchmark for Multimodal Large Language Models},
  author={Fu, Chaoyou and Chen, Peixian and Shen, Yunhang and Qin, Yulei and Zhang, Mengdan and Lin, Xu and Qiu, Zhenyu and Lin, Wei and Yang, Jinrui and Zheng, Xiawu and Li, Ke and Sun, Xing and Ji, Rongrong},
  journal={arXiv preprint arXiv:2306.13394},
  year={2023}
}

@article{mmbench,
  title={MMBench: Is Your Multi-modal Model an All-around Player?},
  author={Liu, Yuan and Duan, Haodong and Zhang, Yuanhan and Li, Bo and Zhang, Songyang and Zhao, Wangbo and Yuan, Yike and Wang, Jiaqi and He, Conghui and Liu, Ziwei and others},
  journal={arXiv preprint arXiv:2307.06281},
  year={2023}
}

@article{seed,
  title={Seed-bench: Benchmarking multimodal llms with generative comprehension},
  author={Li, Bohao and Wang, Rui and Wang, Guangzhi and Ge, Yuying and Ge, Yixiao and Shan, Ying},
  journal={arXiv preprint arXiv:2307.16125},
  year={2023}
}

@article{video-llava,
  title={Video-llava: Learning united visual representation by alignment before projection},
  author={Lin, Bin and Ye, Yang and Zhu, Bin and Cui, Jiaxi and Ning, Munan and Jin, Peng and Yuan, Li},
  journal={arXiv preprint arXiv:2311.10122},
  year={2023}
}

@article{shortv,
  title={ShortV: Efficient Multimodal Large Language Models by Freezing Visual Tokens in Ineffective Layers},
  author={Yuan, Qianhao and Zhang, Qingyu and Liu, Yanjiang and Chen, Jiawei and Lu, Yaojie and Lin, Hongyu and Zheng, Jia and Han, Xianpei and Sun, Le},
  journal={arXiv preprint arXiv:2504.00502},
  year={2025}
}

@inproceedings{goyal2017making,
  title={Making the v in vqa matter: Elevating the role of image understanding in visual question answering},
  author={Goyal, Yash and Khot, Tejas and Summers-Stay, Douglas and Batra, Dhruv and Parikh, Devi},
  booktitle={Proceedings of the IEEE conference on computer vision and pattern recognition},
  pages={6904--6913},
  year={2017}
}

@article{video-chatgpt,
  title={Video-chatgpt: Towards detailed video understanding via large vision and language models},
  author={Maaz, Muhammad and Rasheed, Hanoona and Khan, Salman and Khan, Fahad Shahbaz},
  journal={arXiv preprint arXiv:2306.05424},
  year={2023}
}

@misc{llava-1.5,
      title={Improved Baselines with Visual Instruction Tuning}, 
      author={Haotian Liu and Chunyuan Li and Yuheng Li and Yong Jae Lee},
      year={2024},
      eprint={2310.03744},
      archivePrefix={arXiv},
      primaryClass={cs.CV},
      url={https://arxiv.org/abs/2310.03744}, 
}

@misc{gpt4v,
 author = {OpenAI},
 title = {GPT-4V(ision) System Card},
 year = {2024}
}

@article{gemini,
  title={Gemini: a family of highly capable multimodal models},
  author={{Gemini Team}},
  journal={arXiv preprint arXiv:2312.11805},
  year={2023}
}

@article{BLIP2,
  title={BLIP-2: Bootstrapping Language-Image Pre-training with Frozen Image Encoders and Large Language Models},
  author={Junnan Li and Dongxu Li and Silvio Savarese and Steven Hoi},
  journal={ArXiv},
  year={2023},
  volume={abs/2301.12597}
}

@article{MiniGPT4,
  title={MiniGPT-4: Enhancing Vision-Language Understanding with Advanced Large Language Models},
  author={Deyao Zhu and Jun Chen and Xiaoqian Shen and Xiang Li and Mohamed Elhoseiny},
  journal={ArXiv},
  year={2023},
  volume={abs/2304.10592},
  url={https://api.semanticscholar.org/CorpusID:258291930}
}

@article{liu2024visual,
  title={Visual instruction tuning},
  author={Liu, Haotian and Li, Chunyuan and Wu, Qingyang and Lee, Yong Jae},
  journal={Advances in neural information processing systems},
  volume={36},
  year={2024}
}

@misc{liu2024llavanext,
    title={LLaVA-NeXT: Improved reasoning, OCR, and world knowledge},
    url={https://llava-vl.github.io/blog/2024-01-30-llava-next/},
    author={Liu, Haotian and Li, Chunyuan and Li, Yuheng and Li, Bo and Zhang, Yuanhan and Shen, Sheng and Lee, Yong Jae},
    month={January},
    year={2024}
}

@inproceedings{singh2019towards,
  title={Towards vqa models that can read},
  author={Singh, Amanpreet and Natarajan, Vivek and Shah, Meet and Jiang, Yu and Chen, Xinlei and Batra, Dhruv and Parikh, Devi and Rohrbach, Marcus},
  booktitle={Proceedings of the IEEE/CVF conference on computer vision and pattern recognition},
  pages={8317--8326},
  year={2019}
}
}

\end{document}